# grapheme-kit: Grapheme-Level Metrics and Text Processing for Multilingual NLP

**Izzath Nisfer[1,*], Ashini Kavindya[1,*], Ovindu Atukorala[1], Purushoth Velayuthan[2], Menan Velayuthan[3]**

[1]Department of Computer Science & Engineering, University of Moratuwa, Sri Lanka
[2]Computer Science and Engineering, University of Nevada, Nevada, USA
[3]Department of Information & Computing Sciences, Utrecht University, Utrecht, The Netherlands

[*]Equal contribution

**Correspondence:** m.velayuthan@uu.nl

## Abstract

Existing lexical distance, similarity, and evaluation metrics operate on Unicode code points, which can misrepresent errors in writing systems where a single grapheme is represented by multiple Unicode code points. We introduce grapheme-kit, an open-source Python library that extends these metrics to operate on grapheme clusters instead. The library also provides improved grapheme processing for Tamil and Sinhala, including accurate grapheme cluster identification and grapheme composition/decomposition utilities. Through an OCR case study, we demonstrate that grapheme-level metrics provide a more faithful evaluation of complex scripts.

PyPI Code Demo Docs Video

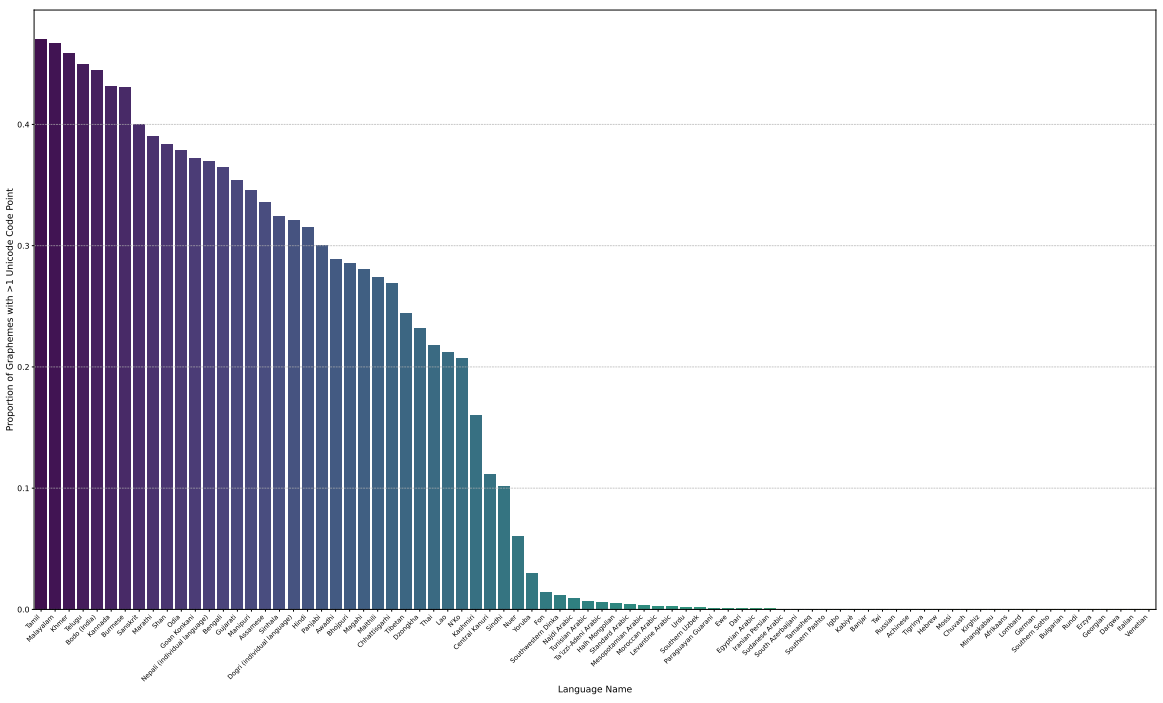

Figure 1: Proportion of grapheme clusters composed of multiple Unicode code points across FLORES+ (Gordeev et al., 2024) dataset. Languages are ordered by the computed proportion in descending order; only the first 80 are displayed.

## 1 Introduction

*"New directions in science are launched by new tools much more often than by new concepts."*

— *Freeman Dyson*

Modern text-based Natural Language Processing (NLP) systems operate on Unicode code points. While this representation works well for many writing systems, it becomes a major limitation for complex-script languages, where a single user-perceived character is often represented by multiple Unicode code points (Petrov et al., 2023; Velayuthan and Sarveswaran, 2025). As shown in Figure 1, a substantial number of languages exhibit a high proportion of graphemes composed of multiple Unicode code points. This disparity between writing systems that predominantly represent a character using a single Unicode code point and those that require multiple Unicode code points introduces additional computational overhead and inefficiencies in NLP systems (Petrov et al., 2023). Grapheme clusters (referred to as *graphemes* hereafter) address this issue by grouping the Unicode code points that together form a single user-perceived character and treating them as a single unit.

While graphemes are extensively used in Automatic Speech Recognition (ASR) (Bunzeck et al., 2024; Dong et al., 2022; Route et al., 2019), Optical Character Recognition (OCR) (Hasan et al., 2025; Elkhayati et al., 2022), and even tokenization (Velayuthan and Sarveswaran, 2025), there remains a lack of comprehensive tooling for grapheme-level text processing and evaluation. To address this, we introduce grapheme-kit, a Python library for grapheme-level text processing and evaluation.

We extend key lexical metrics used in modern NLP systems to operate on graphemes instead of Unicode code points as the fundamental lexical unit. We group these metrics into three categories: (i) distance metrics, (ii) similarity metrics, and (iii) evaluation metrics. Figure 2 provides a complete overview of the features supported by our library.

Current libraries for identifying grapheme clusters fail to correctly identify certain grapheme clusters in Tamil and Sinhala (see Table 2 for examples).

We find that these errors primarily arise from the handling of Zero Width Joiner (ZWJ) and Zero Width Non-Joiner (ZWNJ) characters. Our library rectifies these issues for both languages, enabling correct grapheme cluster identification. We also observe similar issues in other Indic scripts, and our library is designed to be easily extended by language experts to support additional languages.

Vowel–consonant decomposition plays an important role in fine-grained linguistic analysis (Ansary et al., 2024). To support such analyses, our library provides utilities to decompose Tamil and Sinhala graphemes into their constituent consonant and vowel components, as well as reconstruct graphemes from these components. The same design can be readily extended to support other Indic scripts.

Our core contributions are as follows:

- We introduce `grapheme-kit`, an open-source Python library that provides a comprehensive toolkit for grapheme-aware text processing and evaluation across languages.
- We extend key lexical metrics to operate on graphemes instead of Unicode code points, supporting distance, similarity, and evaluation metrics.
- We improve grapheme processing for Tamil and Sinhala by correcting grapheme cluster identification and providing utilities for vowel–consonant decomposition, while designing the library to be easily extended to other Indic scripts.
- We make the library[1] publicly available through PyPI, accompanied by comprehensive documentation[2] and an interactive live demo[3] to facilitate adoption by the research community.

## 2 Related work

### 2.1 Graphemes

Graphemes are the fundamental units of writing (Meletis, 2019). However, modern NLP systems predominantly operate on Unicode code points and subword tokenization algorithms such as Byte Pair Encoding (BPE) and Unigram language models (Sennrich et al., 2016; Kudo, 2018), which often fail to preserve the orthographic structure of complex scripts.

In scripts such as Tamil and Sinhala, a single grapheme is typically represented by multiple Unicode code points comprising base consonants, vowel signs, and other combining characters. Consequently, conventional tokenization algorithms frequently split graphemes into incomplete units, resulting in fragmented linguistic representations and longer token sequences (Petrov et al., 2023; Velayuthan and Sarveswaran, 2025). Recent work has therefore proposed grapheme-aware tokenization methods such as Grapheme Pair Encoding (GPE) and Constrained Byte Pair Encoding (CBPE) (Velayuthan and Sarveswaran, 2025; Brahma et al., 2025).

Existing grapheme segmentation libraries provide only partial support for Indic scripts and fail to correctly handle several orthographic patterns in Tamil and Sinhala. We compare these libraries in Table 1.

### 2.2 Evaluation metrics

Automatic evaluation metrics are fundamental to measuring the quality of NLP systems (Celikyilmaz et al., 2021). Many lexical metrics are built upon string comparison techniques such as Hamming distance and Levenshtein distance (Hamming, 1950; Levenshtein, 1965).

Character Error Rate (CER) extends Levenshtein distance and serves as the standard evaluation metric for Automatic Speech Recognition (ASR) and text normalization (K et al., 2024; Ridoy et al., 2025; Bekarystankyzy et al., 2024; Ghimire et al., 2025). For machine translation and text generation, BLEU measures word-level $n$-gram overlap, while chrF and chrF++ operate on character $n$-grams and correlate well with human judgments, particularly for morphologically rich and low-resource languages (Papineni et al., 2002; Popović, 2015; Popovic, 2016).

Despite their widespread adoption, these metrics fundamentally operate on Unicode code points, characters, or words as their lexical units. To the best of our knowledge, there exists no unified toolkit that extends these commonly used lexical metrics to operate directly on grapheme clusters.

### 2.3 Sinhala and Tamil Languages

Sinhala is an Indo-Aryan language spoken by over 17 million people in Sri Lanka, while Tamil is a Dravidian language spoken by more than 78 million people across South Asia and the global Tamil diaspora (Krishnamurti, 2003; De Silva, 2019). Despite their large speaker populations, both remain low-resource languages in NLP research (Ranathunga and de Silva, 2022; Tissera and Saadhiq, 2025). Both writing systems are derived from the ancient Brahmi script and belong to the abugida family, where graphemes are formed by combining base consonants with vowel signs and other combining characters (Kunchukuttan et al., 2021). Consequently, a single user-perceived character is often represented by multiple Unicode code points, making Unicode-based tokenization and evaluation less suitable for these languages and motivating grapheme-aware processing (Petrov et al., 2023; Velayuthan and Sarveswaran, 2025; Fernando and Ranathunga, 2025).

[1] https://pypi.org/project/grapheme-kit/
[2] https://grapheme-kit-docs.pages.dev/
[3] https://grapheme-kit.pages.dev/

| Feature / Limitation | Grapheme | U-grapheme | Indic NLP | Regex | grapheme-kit |
|---|---|---|---|---|---|
| Sinhala support | 'ක්\u200d', 'ර', 'ම', 'ය' | 'ක්\u200d', 'ර', 'ම', 'ය' | 'ක්', '\u200d', 'ර', 'ම', 'ය' | 'ක්\u200d', 'ර', 'ම', 'ය' | 'ක්\u200dර', 'ම', 'ය' |
| Tamil support | 'ஸ்', 'ரீ', ' ', 'ல', 'ங்', 'கா' | 'ஸ்', 'ரீ', ' ', 'ல', 'ங்', 'கா' | 'ஸ்ரீ', 'ல', 'ங்கா' | 'ஸ்', 'ரீ', ' ', 'ல', 'ங்', 'கா' | 'ஸ்ரீ', ' ', 'ல', 'ங்', 'கா' |

Table 1: Comparison of libraries for grapheme-level segmentation in Sinhala and Tamil

## 3 System Overview

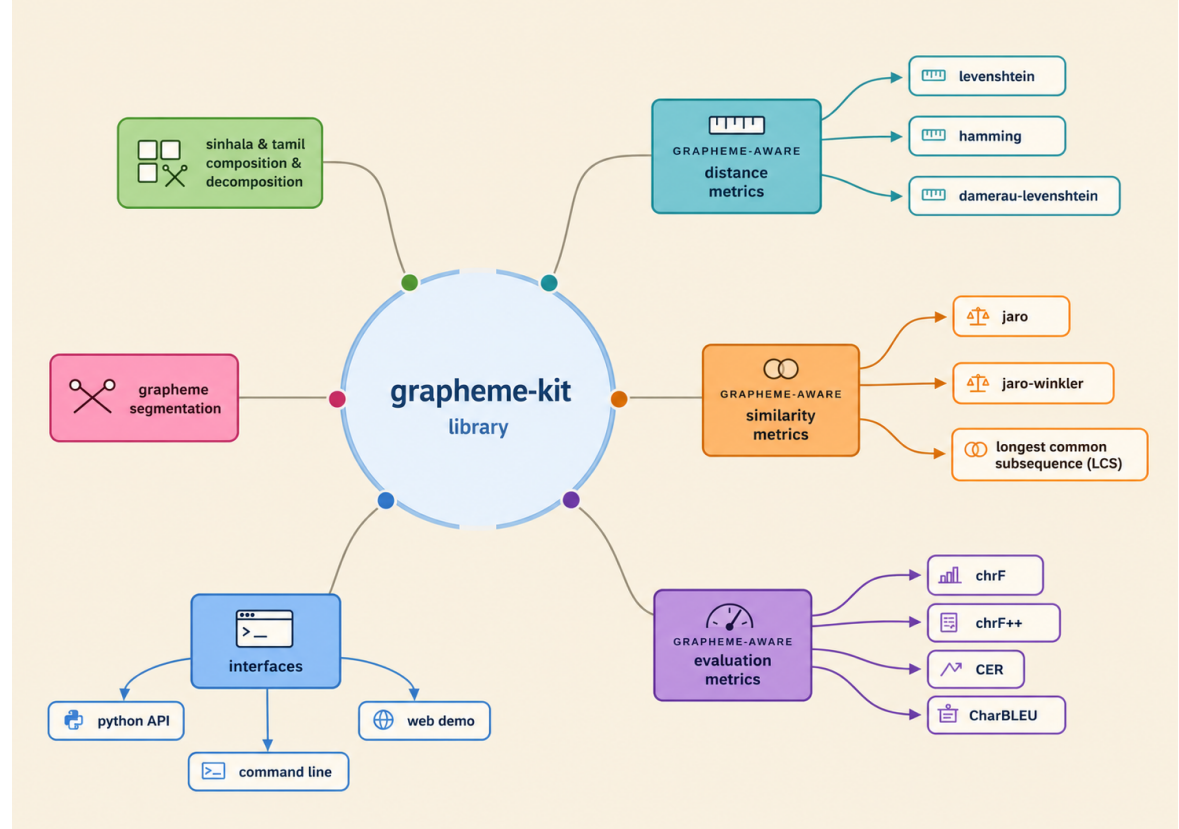

Figure 2: Overview of the grapheme-kit library architecture. Inspiration for the diagram was taken from (Suzgun et al., 2024).

### 3.1 Architecture Overview

Raw input text first passes through a normalization stage that resolves Unicode level inconsistencies and language specific encoding irregularities. The normalized text is then segmented into grapheme clusters using graphemizer. Then these clusters are used to implement grapheme aware string distances, grapheme level evaluation metrics, and composition and decomposition of clusters into their constituent base consonant and vowel components.An overview of this workflow is illustrated in Figure 2.

### 3.2 Graphemizer

Graphemizer builds on top of grapheme libarary. When code points go through graphemizer, it identifies specific cases where grapheme library fails and gives exact visual units in scripts(refer Table 2).

In Sinhala there are some complex consonant clusters by joining a sequence of consonants through ZWJ characters, each carrying its own virama and vowel modifiers. Grapheme library treats a ZWJ as a boundary between two separate clusters .So a single visually and linguistically unified character gets split into multiple grapheme clusters. Graphemizer detects ZWJ linked sequences and recursively folds every consonant, virama, and vowel modifier joined by ZWJ into a single grapheme cluster regardless of how many ZWJs are involved in the chain.

### 3.3 Grapheme Level Metrics

Grapheme level metrics evaluate and compare strings using grapheme clusters rather than individual Unicode code points

For example, consider the hindi words किताब(book) and कताब. Their grapheme cluster representations are कि, ता, ब and क, ता, ब.

Although the Unicode representations differ by several individual code points, the grapheme level representation groups each user perceived character into a single unit. metrics operate on those visible characters perceived by readers.

This library implements several distances, similarity metrics, and evaluation metrics in grapheme level.

- **Distance Metrics:** Levenshtein, Hamming, Damerau-Levenshtein

| | Sinhala | Tamil | English |
|---|---|---|---|
| Unicode codepoints | 'ශ', '්', '\u200d' , 'ර', 'ී', ' ', 'ල', 'ං', 'ක', 'ා', 'ව' | 'ஸ', '்', 'ர', 'ீ', ' ', 'ல', 'ங', '்', 'க', 'ா' | 'S', 'r', 'i', ' ', 'l', 'a', 'n', 'k', 'a' |
| Grapheme Clusters | 'ශ්\u200d', 'රී', ' ', 'ලං', 'කා', 'ව' | 'ஸ்', 'ரீ', ' ', 'ல', 'ங்', 'கா' | 'S', 'r', 'i', ' ', 'l', 'a', 'n', 'k', 'a' |
| Grapheme-kit Clusters | 'ශ්\u200dරී', ' ', 'ලං', 'කා', 'ව' | 'ஸ்ரீ', ' ', 'ல', 'ங்', 'கா' | 'S', 'r', 'i', ' ', 'l', 'a', 'n', 'k', 'a' |

Table 2: segmentation on the phrase "Sri lanka" translated into Tamil and Sinhala.

- **Similarity Metrics:** Jaro, Jaro-Winkler, Longest Common Subsequence (LCS)
- **Evaluation Metrics:** chrF, chrF++, Character Error Rate (CER), CharBLEU

### 3.4 Grapheme Composition and Decomposition

To support editing and analysis at a sub-grapheme level, grapheme-kit provides inverse *composition* and *decomposition* operations that convert between a full grapheme cluster and its constituent base consonant and vowel components.

**Composition.** Given a consonant carrying a virama (e.g. ம் or ක්) and a target independent vowel, the composer retrieves the corresponding dependent vowel sign and combines it with the base consonant. When the target vowel is the inherent vowel (அ or අ), the virama is removed and the bare consonant is returned. For example, ம் combined with இ yields மி, while ක් combined with ඉ yields කි. Likewise, க் combined with அ produces க, and ක් combined with අ produces ක.

For example:

- ந்ஆர்இ → நாரி
- ව්ඓද්ය්අ → වෛද්‍ය

**Decomposition.** Decomposition reverses this process by representing each grapheme as a base consonant with an explicit virama followed by its corresponding independent vowel. Independent vowels decompose to themselves paired with an empty vowel component. For example, the Tamil grapheme மி decomposes to ம் and இ, while the Sinhala grapheme කි decomposes to ක් and ඉ. Similarly, bare consonants are interpreted as carrying the inherent vowel, so க decomposes to க் and அ, and ක decomposes to ක් and අ.

For example:

- நிலுவை → ந்இல்உவ்ஐ
- ජ්‍යෝතිෂ්‍ය → ජ්ය්ඕත්ඉෂ්ය්අ

### 3.5 Interfaces

`grapheme-kit` is exposed through two interfaces that share the same underlying implementation: a Python API for programmatic use inside data and training pipelines, and a command line interface (CLI) for quick, script-free inspection and batch processing from the shell. Every operation described above—segmentation, distance and evaluation metrics, and composition/decomposition—is available identically through both.

**Python API.** Each operation is a small, composable function or class. Grapheme segmentation is provided by the `Graphemizer` class, whose `graphemes` attribute yields the grapheme-kit clusters, while distances, evaluation metrics, and composition/decomposition are plain functions that operate directly on strings. For further examples, see the package documentation.

```
from grapheme_kit import Graphemizer

g = Graphemizer("مُعَلِّمٌ")
g.graphemes    # ['مٌ', 'لِّ', 'عَ', 'مُ']
len(g)    # 4
```

**Command line interface.** The same operations are available from the shell through the `grapheme-kit` command (aliased as `gkit`), with one subcommand per operation. This lets grapheme-aware segmentation and evaluation be scripted or piped without writing any Python. For further details, refer to the CLI documentation.

```
$ gkit graphemize "किताब" --count
graphemes:   3
code points: 5

$ gkit evaluate "ကျွန်ုပ်"
      "ကျွန်ုပ" --metric chrf
chrF (grapheme): 38.8889
```

## 4 Case-study evaluations

### 4.1 Case Study: Grapheme-Aware Evaluation

To demonstrate the effect of grapheme-aware evaluation, we compare grapheme-level variants of Character Error Rate (grapheme-CER) and chrF++ (grapheme-chrF++) against their standard codepoint-level counterparts on an OCR task. We use OCR as a representative case study, as it clearly exposes the measurement bias introduced by Unicode code point-based evaluation. While our experiments are limited to OCR, the proposed metrics are applicable to other tasks that rely on lexical evaluation, such as Automatic Speech Recognition (ASR) and Machine Translation (MT). We evaluate across 12 languages spanning Latin controls (Dutch and French), Arabic, and scripts with substantial Unicode decomposition (Hindi, Bengali, Tamil, Telugu, Malayalam, Sinhala, Sanskrit, Dzongkha, and Burmese). OCR outputs are obtained by rendering FLORES+ (Gordeev et al., 2024) sentences using Noto fonts and recognizing them with Tesseract.

**Script complexity.** We quantify the degree of Unicode decomposition using the *grapheme inflation ratio*, defined as the average ratio between grapheme clusters and Unicode code points per reference sentence. A ratio close to 1.0 indicates that a visual character is typically represented by a single Unicode code point, whereas lower values indicate that graphemes are composed of multiple code points. As shown in Table 3, Telugu, Burmese, Malayalam, Tamil, Hindi, Sanskrit, Bengali, and Sinhala exhibit substantial decomposition (0.59–0.65), while Arabic (0.99) behaves similarly to Latin scripts and Dzongkha (0.76) lies in between. Such scripts are expected to benefit most from grapheme-level metrics, as a single grapheme spanning multiple Unicode code points may otherwise incur multiple edit operations under codepoint-level evaluation.

| Language | Inflation Ratio |
|---|---|
| Telugu | 0.59 |
| Burmese | 0.60 |
| Malayalam | 0.60 |
| Tamil | 0.61 |
| Hindi | 0.62 |
| Sanskrit | 0.62 |
| Bengali | 0.64 |
| Sinhala | 0.65 |
| Dzongkha | 0.76 |
| Arabic | 0.99 |
| Dutch | 1.00 |
| French | 1.00 |

Table 3: Grapheme inflation ratio across languages. Lower values indicate greater Unicode decomposition.

**OCR: Grapheme-level metrics correct a substantial measurement bias.** OCR is the setting where this bias is most apparent, as OCR errors often arise from misrecognizing a single component within an otherwise correct grapheme cluster (e.g., a dependent vowel sign). Under codepoint-level evaluation, such errors are disproportionately penalized despite affecting only a single user-perceived character. Table 4 reports CER and chrF++ (word bigrams + character $n$-grams, with $n$=3 for the grapheme variant) and shows consistent improvements under grapheme-aware evaluation. As expected, languages where grapheme clustering has little effect (Arabic and the Latin controls) show virtually no change. In contrast, Burmese and Dzongkha exhibit a small decline, as the OCR system frequently fails to produce text resembling the reference, leaving little component-level information for grapheme-aware evaluation to recover.

| Language | char-CER | g-CER | chrF++ | g-chrF++ | ΔchrF++ |
|---|---|---|---|---|---|
| Tamil | 5.48 | 0.62 | 74.05 | 98.99 | +24.94 |
| Sinhala | 3.80 | 0.92 | 86.45 | 98.88 | +12.43 |
| Malayalam | 11.35 | 9.68 | 72.55 | 78.14 | +5.59 |
| Bengali | 6.36 | 4.59 | 89.27 | 93.86 | +4.59 |
| Telugu | 3.38 | 3.09 | 90.76 | 95.30 | +4.54 |
| Sanskrit | 3.53 | 3.24 | 87.59 | 92.01 | +4.42 |
| Hindi | 0.70 | 0.71 | 98.30 | 98.41 | +0.11 |
| Dutch | 0.13 | 0.13 | 99.52 | 99.52 | +0.00 |
| French | 0.58 | 0.58 | 98.51 | 98.51 | +0.00 |
| Arabic | 82.96 | 83.32 | 23.46 | 23.38 | −0.08 |

Table 4: OCR results: standard vs. grapheme-level metrics (chrF++ at tuned grapheme $n$=3). Languages sorted by improvement.

### 4.2 Controlled Perturbation Analysis

The case study above evaluates metrics on real model outputs, where script complexity, model quality, and error types are inherently coupled. To isolate the effect of grapheme-level evaluation, we construct a controlled perturbation study following prior work on metric diagnostics, which emphasizes evaluating metrics under targeted edits of known types rather than relying only on aggregate correlations (Karpinska et al., 2022). We apply a set of controlled perturbations (ref Appendix B.1) directly to the same FLORES+ reference sentences used above (1,012 sentences per language across all 12 languages), allowing the same edit operation to be compared across scripts with varying degrees of grapheme inflation. The perturbations include word-level modifications (repetition, reordering, and shuffling), a realistic misspelling through vowel-sign substitution, two character-deletion variants, and sanity-check baselines (empty hypothesis, unrelated sentence, and reference-as-hypothesis).

**A minimal-pair comparison isolates the mechanism directly.** We focus on a character deletion perturbation implemented at two different levels:

codepoint-level deletion, where a single Unicode code point is removed (e.g., a dependent vowel sign), and grapheme-level deletion, where an entire grapheme cluster is removed. The former corrupts an otherwise intact grapheme cluster, while the latter removes a complete linguistic unit. Table 5 shows that these two operations produce opposite biases under character-level evaluation. For codepoint-level deletion, grapheme-CER assigns higher error than char-CER (+0.60 points averaged across the ten non-Arabic complex scripts), reflecting that the corrupted grapheme is a visible error despite involving only a single codepoint change. Conversely, for grapheme-level deletion, grapheme-CER assigns lower error ($-0.31$), as it treats the removal of a complete grapheme as a single edit, while char-CER may count multiple codepoint edits. Latin controls (Dutch and French) show no difference, as grapheme clustering is effectively a no-op. These results demonstrate that the observed differences are specific to scripts with grapheme inflation and arise from the choice of evaluation unit rather than the edit-distance computation itself.

| Perturbation | CER | g-CER | ΔCER | chrF | g-chrF (Δ) |
|---|---|---|---|---|---|
| Misspelled word (25) | 1.22 | 1.12 | −0.10 | 97.89 | 97.31 (−0.58) |
| Delete one codepoint (26a) | 0.82 | 1.42 | +0.60 | 98.32 | 97.19 (−1.14) |
| Delete one grapheme (26b) | 1.42 | 1.12 | −0.31 | 97.70 | 97.59 (−0.12) |

Table 5: Controlled perturbations (mean over 12 languages; chrF at tuned grapheme $n$=3). ΔCER = grapheme − standard; ΔchrF = grapheme − standard.

**chrF: consistent for corruption, an open question for clean substitution.** Grapheme-chrF agrees with grapheme-CER on the codepoint-deletion perturbation (26a), assigning lower scores to the corrupted outputs than standard chrF. However, for the clean-substitution perturbations (25, 26b), grapheme-chrF does not follow the more forgiving behaviour observed in grapheme-CER and instead assigns lower scores than standard chrF. This difference arises from the underlying formulation of the metrics. Grapheme-CER benefits from treating a multi-codepoint grapheme as a single edit unit, reducing the edit penalty for clean substitutions. In contrast, chrF measures $n$-gram overlap rather than discrete edit operations; therefore, a changed grapheme can disrupt similar numbers of $n$-gram matches regardless of whether it spans one or multiple codepoints. Since grapheme-level $n$-grams operate over a smaller unit space, localized changes can represent a larger fraction of the overall $n$-gram overlap, causing grapheme-chrF to move more than standard chrF. We leave further investigation of this behaviour and potential normalization strategies for grapheme-chrF as future work.

## 5 Conclusion

We present `grapheme-kit`, an easy-to-use Python library for grapheme-level text processing and evaluation. We extend widely used lexical metrics to operate on grapheme clusters and provide language-aware grapheme processing utilities for complex scripts. Through multilingual case studies across machine translation, automatic speech recognition, and optical character recognition, we demonstrate the benefits of grapheme-level evaluation for scripts with substantial Unicode decomposition. We hope that `grapheme-kit` provides an accessible foundation for developing and evaluating multilingual NLP systems using linguistically meaningful text units.

## Limitations

We have currently identified two limitations in our library. First, the accuracy of our evaluation metrics depends on the underlying grapheme segmentation library. Second, the composition and decomposition features are available for only two languages at present. We are actively working with a group of excellent volunteers to extend these features to more Indic languages.

## Acknowledgments

We, the authors, would like to thank the developers and maintainers of the open-source libraries that supported our implementation, specifically the `grapheme` library[4] for Unicode grapheme cluster segmentation, the `textdistance` library[5] for string distance and similarity computations, and the `sacrebleu` library[6] for evaluation metrics.

[4] https://github.com/itcarroll/grapheme
[5] https://github.com/orsinium/textdistance
[6] https://github.com/mjpost/sacrebleu

# A Text Pre-processing

## A.1 Normalization

`grapheme-kit` provides a language aware normalization module for Tamil and Sinhala to ensure that visually and linguistically equivalent text is represented in a canonical form before further processing. The normalization pipeline consists of two stages.

**Unicode Normalization** The input text is first transformed into Unicode Normalization Form C (NFC), which applies canonical composition to combine decomposed Unicode sequences into their standard composed representation. This step guarantees a consistent Unicode encoding for equivalent character sequences.

**Language specific Normalization.** After NFC normalization, a set of rule based transformations is applied to correct script specific inconsistencies. For Tamil, these rules normalize alternative vowel sign sequences, remove Zero Width Joiner (ZWJ) and Zero Width Non-Joiner (ZWNJ) characters, correct reversed vowel ordering, and replace non standard character combinations with their canonical forms. For example, the sequence ொ is normalized to ொ, and ோ is normalized to ோ.

Similarly, Sinhala normalization applies a confusion set mapping to correct frequently occurring ordering and composition errors involving dependent vowel signs and virama characters. For example, the non-canonical sequence ේා is normalized to ෝ, and ෟෙ is normalized to ෞ. These transformations ensure that equivalent grapheme clusters are represented consistently, improving the reliability of grapheme segmentation and subsequent evaluation metrics.

# B Evaluations

## B.1 Controlled perturbation

We adapt a subset of Karpinska et al. (2022)'s (DEMETR) perturbation taxonomy. Most of the perturbations are word-level edits, where the codepoint-versus-grapheme distinction is immaterial, since word boundaries always contain whole grapheme clusters. We focus the main analysis on the three that vary edit granularity directly:

- **25 - Misspelled word:** one grapheme cluster's vowel is substituted.
- **26a - Delete one codepoint:** corrupts a multi-codepoint cluster.

- **26b - Delete one grapheme cluster:** removes a whole, possibly multi-codepoint grapheme cluster.